%% file: main.tex
\def\year{2022}\relax
\documentclass[letterpaper]{article} 
\usepackage{aaai22}  
\usepackage{times}  
\usepackage{helvet}  
\usepackage{courier}  
\usepackage[hyphens]{url}  
\usepackage{graphicx} 
\usepackage{natbib}  
\usepackage{caption} 
\DeclareCaptionStyle{ruled}{labelfont=normalfont,labelsep=colon,strut=off} 
\usepackage{algorithm}
\usepackage[noend]{algorithmic}
\nocopyright
\usepackage{verbatim}

\usepackage{amsmath, amsthm, amssymb, thmtools}

\usepackage{xspace}
\newcommand{\asec}{$\mathsf{ASEC}$\xspace}

\title{Position Paper:\\
	Online Modeling for Offline Planning}
\author{
    Eyal Weiss and Gal A. Kaminka
}
\affiliations{

	The MAVERICK Group, Bar-Ilan University, Israel\\
    \{weissey, galk\}@cs.biu.ac.il
}

\begin{document}

\maketitle

\begin{abstract}
The definition and representation of planning problems is at the heart of AI planning research. A key part is the representation of action models.
Decades of advances improving declarative action model representations resulted in numerous theoretical advances, and capable, working, domain-independent planners.
However, despite the maturity of the field, AI planning technology is still rarely used outside the research community, suggesting that current representations fail to capture real-world requirements, such as utilizing complex mathematical functions and models learned from data.
We argue that this is because the modeling process is assumed to have taken place and completed prior to the planning process, i.e., \emph{offline modeling} for \emph{offline planning}.
There are several challenges inherent to this approach, including: limited expressiveness of declarative modeling languages; early commitment to modeling choices and computation, that preclude using the most appropriate resolution for each action model---which can only be known \emph{during planning}; and difficulty in \emph{reliably} using non-declarative, learned, models.

We therefore suggest to change the AI planning process, such that is carries out \textit{online modeling in offline planning}, i.e., the use of action models that 
are computed or even generated as part of the planning process, as they are accessed. This generalizes the existing approach (offline modeling).
The proposed definition admits novel planning processes, and we suggest one concrete implementation, demonstrating the approach.
We sketch initial results that were obtained as part of a first attempt to follow this approach by planning with action cost estimators. We conclude by discussing open challenges.

\end{abstract}

\input{introduction}

\input{our_approach}
\input{criteria}

\input{research_status}

\input{challenges_future_work}

\section*{Acknowledgements}
The research was partially funded by ISF Grant \#2306/18 and BSF-NSF grant 2017764. Thanks to K. Ushi. 
E.W. is supported by the Adams Fellowship Program of the Israel Academy of Sciences
and Humanities and by Bar-Ilan University's President Scholarship.

\bibliography{short,weiss_references,aicyber} 
\end{document}

%% file: introduction.tex
\section{Introduction}
The definition and representation of planning problems is at the heart of AI planning research.
Traditionally, planning (really, offline planning, i.e., planning before execution) has been defined as the process of generating a plan or policy
that will take an agent that executes it to a state that achieves (or maximizes) its goals. Implicit in this definition is reliance on an \emph{action model}, created offline (ahead of the planning process). This  allows the process to consider the actions the agent may take at any given state, what their effects will be, and at what costs.

The representation of action models given to the planner is a main topic of research.
Over the years, steady efforts have been made 
to develop action model representations that are both general and expressive using declarative languages~\cite{fox2003pddl2,baier2008beyond, eyerich2013beyond}. 
Optimized open-source software implementations (e.g.,~\cite{helmert-jair2006}), which admit these representations,
can use sophisticated search algorithms~\cite{lipovetzky2017best}, and efficient domain-independent heuristics (see~\cite{ghallab2016automated}),
to solve challenging planning problems in a variety of domains.


However, despite the maturity of the field, AI planning technology is still rarely used outside the research community.
After decades of advances improving action model
representations, they remain disconnected from many real world requirements~\cite{blythe1999decision,mccluskey2003pddl,boddy2003imperfect,rintanen2015impact}. 
In some cases, the effects of actions are too complex to describe declaretively, 
or there may be uncertainty as to the effects or their cost. In some cases,
 models may only be available in black-box form (e.g., when learned~\cite{arora2018review}); this limits their usability in different planners.  Furthermore, some models are computationally expensive to build in advance of the planning, so preparing informative action models may be unfeasible. For instance, declaring the cost of possible 
ground actions using external sources, ahead of the planning process, is practically impossible in real-world domains. 

Semantic attachments, which were first introduced in~\cite{weyhrauch1980prolegomena} and incorporated in PDDL by~\cite{dornhege2009semantic}, is an important concept that offers a partial solution to some of the problems raised above. 
It allows combining external procedures that supplement declarative models by performing condition checks and calculating action effects, thus offering access to modeling of complex mathematical functions and postponing calculations involving them to the planning process.
However, this ignores the run time and of external procedures, and the uncertainty they may represent.
When using data-driven models for planning, one has to address these issues, so as to decrease the computational burden of modeling (while planning),
and increase the reliability of the plan found. 

We argue that the common cause for the above difficulties is that the modeling process is assumed to have taken place
and completed prior to the planning process, i.e., \emph{offline modeling} for \emph{offline planning}.  The current
approach to planning requires that the modeling process, whether carried out manually or by learning, generates a single
set of action models, which the planner later uses during planning.  This raises the following challenges inherent to this approach:

%

\paragraph{Waiting for the perfect PDDL.} To be general, planners must be able to work with action model description languages that capture complex effects and conditions.  Yet despite decades of advances in PDDL, 
that there are still many
cases in which its declarative expressiveness is not sufficient, e.g., because of limited mathematical operator set, or because some
state factors are not easily described by truth-values or even purely numerically.  Of course, we should continually improve and extend
PDDL, but this is a slow process, and even slower in terms of building planners able to support advanced features.
Even today, few if any planners admit the full set of features of the latest PDDL. 

\paragraph{Inability to work with multiple languages.} As the perfect modeling language is still in our future,
existing planners in challenging areas of significant practical value, use hybrid models.  For example, in Task and Motion Planning (TAMP), both a task description must be provided (e.g., in PDDL), as
well as description of geometric and kinematic considerations of motion~\cite{garrett2021integrated}.

\paragraph{Early commitment to modeling choices.} Even supposing a single language is sufficient, one may create alternative models, at different levels of detail or accuracy, for the same action. When the modeling choice is made ahead of the planning process, a commitment must be made to the modeling level ahead
of the planning process.  Conservatively, one must choose the most detailed and accurate model, but this often entails significant
planning computations later on.  Thus, for instance, the use of multiple (increasingly accurate) estimators for cost or uncertainty is impossible.  

\paragraph{Early commitment to modeling computation.}  The last point also ties to this one, which is the commitment ahead of planning to a single model per action, regardless of the run time it consumes.
In a fully declarative setting, this means conducting all modeling computations in advance, which might be impracticable.
For example, if the model needs the time it takes to drive a specific road, then this information must be determined ahead of the planning (e.g., by accessing Google Maps).  Even a 50ms query is a long process by computational standards, and certainly when
every such possible ground action must be evaluated ahead of planning.
Even when semantic attachments are used, early commitment to the model might incur significant overhead in case lighter models can be used for some actions (possibly offering reduced, but sufficient, accuracy).

\paragraph{Difficulty in using non-declarative, learned, models.}  
Models learned from data almost always include some form of uncertainty.
Moreover, the accuracy of learned models often depend on the run time allocated for them (e.g., an anytime procedure that continually improves until it is stopped).
When the models are fixed in advance of planning, the planner has limited ability to affect the overall uncertainty of the plan it returns, thereby reducing its effectiveness to robustly cope with learned models.

\vspace{8pt}\noindent We therefore suggest to change the AI planning process, such that is carries out \textit{online modeling} in offline planning, i.e., the use of action models that
are computed or even generated as part of the planning process, as they are dynamically accessed. This generalizes the existing approach (offline modeling). 
Presently, there are two major lines of research taking this approach: \textit{planning with simulators}~\cite{frances17}, and domain-specific attempts, often within TAMP.
While these are good examples and successful in some applications, they do not offer a full solution to the gap in problem modeling.  
Indeed, \textit{planning with simulators} 
sacrifices much mathematical structure---inherent in declarative action models---rendering many known heuristics inapplicable. TAMP is domain-specific, and thus does not offer a high level of generality.
More generally, there are theoretical questions that arise from this changed perspective on planning, which have not been addressed.
We propose to re-formulate the definition of planning, to allow online modeling. Rather than relying on a set of pre-computed action models, the planner should accept a set of action model estimators which are called to dynamically generate action models as needed.  Multiple models can even be generated for the same action, thus allowing reasoning about models at different levels of resolution (and possibly computational cost). 

The formulation admits existing efforts as special cases. Offline modeling is a special case where there is a
single estimator for every action, and it uses a declarative description.
Planning with semantic attachments is a special case where \textit{modeling computation} is conducted online, but modeling choices are pre-fixed. 
Planning with simulators is a special case where \textit{the effects} of actions are generated by calling simulators.  For hybrid TAMP planners, the applicability of a motion (preconditions), or its duration (cost) are computed by calling a local motion planner. 

The proposed definition admits novel planning processes.  For example, under this formulation the initial declarative model may be viewed as a first-order approximation of the correct model, where each application of an estimator improves the approximation.
This closely follows Rao's vision of model-lite planning~\cite{kambhampati2007model} for working with approximate models in a systemic manner. 
It also resembles the work on planning and acting~\cite{ghallab2016automated} which aims for continual refinement of deliberative models during the execution of the plans (i.e., online planning); in contrast, we focus on offline planning.

We briefly report here on 
our work to modify standard optimization criteria to take into account target plan accuracy, in order to  trade-off modeling uncertainty against computation time. The planner calls on more computationally expensive (and more accurate) estimates for the same action, but only as needed.
By explicitly considering model uncertainty and setting a target plan accuracy, we offer a systematic approach to increase reliability, as the cumulative effects of uncertainty (typically induced by data-driven models) are bounded, while reducing unnecessary computational (modeling) overhead, which is crucial for scaling to large problems.
We sketch initial results that were obtained as part of a first attempt to follow this approach by planning with action cost estimators. We conclude by discussing open challenges.


%% file: our_approach.tex
\section{Online Action Models}

The commonly-accepted definition for planning problems is a tuple $P=(\mathcal{I},\mathcal{G},\mathcal{A})$, whose components represent the initial state(s), the goal states,
and the set of action models, respectively.  Each action model $a\in \mathcal{A}$ has associated preconditions, $PRE(a)$, effects $EFF(a)$, 
cost $c(a)$, and sometimes duration or effect probabilities. In other words, every action model
in $a$ is a complex object, which can be queried for information about the action.  Commonly, the answers to the queries remain static throughout the planning process.

We propose to generalize the approach to offline planning, so that it allows the answers to the queries to dynamically change during the
planning process. The proposed modification in definition does not, at a high level, change the tuple components. Instead, the action model objects
$a\in \mathcal{A}$ are considered to be dynamic estimators, rather than static. But this profoundly changes our conception of planning.

We point out several appealing properties of the generalized approach. First, any kind of estimator can be used, so there are no restrictions on the type of data being processed during planning, nor on the mathematical operators being utilized, and in particular estimators can be black-box.
Second, action model estimators can be based on the outcome of domain-specific planning systems (such as motion planners), so that a form of hierarchical planning may take place, \emph{allowing the flow of bottom up information during planning}.
Third, mixed declarative-procedural problem formulations are supported, allowing declarative-based state-of-the-art domain-independent heuristics to retain relevance.
Lastly, model uncertainty can be systematically controlled to meet target plan accuracy, while offering significant potential savings on redundant modeling time.

We distinguish families of action model estimators. Two of specific interest are those that deal with estimation of the symbolic structure of the model, i.e., the preconditions and effects (e.g., as in planning with simulations), and those that estimate the numeric parameters of the actions,
such as costs, duration, or probabilities of effects.  Some TAMP planners take advantage of both types.  Other planners, e.g., those that interleave calls to external sources of information as part of the planning, are equally described in this definition.

We focus here on one specific---novel---type of planning problems which can be described as online modeling for offline planning.
In these, the planner begins with an initial action model, and then utilizes external sources of information for model completion, ad hoc. 
Concretely, we focus on the following case:

\begin{itemize}
	\item The problem's symbolic \emph{structure} is described using declarative action models with structural preconditions and effects (e.g. via predicates).  These remain static.
	\item The \emph{numeric} model parameters are estimated online, with increasing accuracy, by letting the planner call estimators that provide information about their values.
	\item The planner is given an acceptable accuracy for the sought-after plan, allowing the planner to trade-off accuracy vs. computation time.
\end{itemize}

To make this explicit, we add a component to the the planning problem definition. In addition to the initial state, goal and the set of action models, we add $\Theta$, a set of action model parameter \emph{estimators}, so that calling such an estimator with an action model returns an updated model. The planning problem explicitly denotes this as a tuple $P=(\mathcal{I},\mathcal{G},\mathcal{A}, \Theta)$.


The immediate implication of the suggested approach is an enhanced ability to represent and solve planning problems, as clearly every declarative representation (that so far was constructed prior to planning) can be completed incrementally by the planner, given appropriate external modules.
The price paid is increased planning time, due to additional computational effort spent on refining the model.
This trade-off is typical for problem generalization, as it entails solving a harder problem.
The main challenge that arises is thus to develop computationally efficient planners, able to balance resource allocation between search effort and model refinement effort.  We discuss this in the next section.

%% file: criteria.tex
\section{When are Dynamic Models Preferable?}


We focus on automated acquisition of action model parameters, where the structural elements defining the problem are constructed offline.
Our target scenario is characterized by problems with structure that may be well described by a factored representation (namely using action templates), but where every ground action potentially has a unique set of parameters (e.g., cost, duration, probabilities etc.) that is not easily defined using a factored representation.
Hence, obtaining numeric values for each action is done separately, so that every action requires a function call---an invocation of an action model estimator.
We use the term estimator as it generalizes a procedure which simply retrieves values to also account for potential parametric uncertainty.
Additionally, we permit several distinct estimators per action, that provide a range of accuracy levels and running times.

For a given planning problem, we denote by
\begin{equation*}
\mathcal{A}=\{a_1, \dots ,a_n\}, \Theta=\{ \theta_1^1, \dots ,\theta_1^{k_1}, \dots, \theta_n^1, \dots ,\theta_n^{k_n} \}
\end{equation*}
the set of ground actions and the set of action model estimators, respectively, where $\theta_i^j$ stands for the $j$th estimator of action $a_i$, and by $t(\theta_i^j)$ the run time of $\theta_i^j$.
W.l.o.g. we assume that $t(\theta_i^q) \le t(\theta_i^m)$ for $q < m$.
Using these notations we can express the run time required for constructing a full offline model:
\begin{equation*}
T_\textit{offline}^\textit{modeling}=\sum_{i=1}^{n} t(\theta_i^{k_i}).
\end{equation*}
Note that if modeling has to be completed before planning then it has to be conservative, i.e., the best estimates available for every action must be used. Otherwise, the potential quality of plans, that can be found by a planner using these models, will decrease.

Examining $T_\textit{offline}^\textit{modeling}$, one can recognize that it may in fact be unknown a priori.
While $n$ is always known, the running times of the estimators might not be known in advance. However, it is reasonable to assume that some knowledge about them is available (such as their order of magnitude), so at least approximating $T_\textit{offline}^\textit{modeling}$ is plausible.
Hence, when $t_\textit{avg}^\textit{best}:=\frac{1}{n}T_\textit{offline}^\textit{modeling}$ is projected to be high, and $n$ is not small, $T_\textit{offline}^\textit{modeling}$ might be very demanding.
To give a concrete example, consider the case that an estimate is obtained via internet access, so network delay dominates estimator run time. In such a case $t_\textit{avg}^\textit{best} \approx 100 $ms is a typical value (see e.g., https://wondernetwork.com/pings).
As for $n$, benchmark problems contain up to $10^6$ ground actions (and even more in extreme cases)~\cite{gnad2019learning}.
Thus, $T_\textit{offline}^\textit{modeling}$ might take dozens of hours.
It is clear that in such cases there is a strong incentive to try to reduce modeling time.

We now turn to express the run time required for modeling in the case of dynamic estimation:
\begin{equation*}
T_\textit{dynamic}^\textit{modeling}=\sum_{i : a_i \in \mathcal{A}_\textit{actual}} t(\theta_i^\textit{actual}),
\end{equation*}
where $\mathcal{A}_\textit{actual}$ is the set of actions that require estimates during planning, and $\theta_i^\textit{actual}$ are the estimators that are applied during planning.
We highlight that these terms are planner-dependent, and they are also clearly unknown prior to planning.
Nevertheless, it is well known that many planning problems could be solved while exploring only a small fraction of the problem state space, and correspondingly considering only a small part of all possible actions.

Therefore, even without knowing the exact value of $|\mathcal{A}_\textit{actual}|$, it is reasonable to expect for a significant reduction in modeling time.
In addition, it might not be necessary to use the most time-consuming estimator for each action encountered during planning, which implies another potential reduction in modeling time if $t_\textit{avg}^\textit{actual}:=\frac{1}{|\mathcal{A}_\textit{actual}|}T_\textit{dynamic}^\textit{modeling} \ll t_\textit{avg}^\textit{best}$.
Overall we may conclude that $T_\textit{dynamic}^\textit{modeling} \leq T_\textit{offline}^\textit{modeling}$, and in many cases it might be that $T_\textit{dynamic}^\textit{modeling} \ll T_\textit{offline}^\textit{modeling}$.

So far we have seen that when considering modeling time alone it is always favorable to use dynamic estimation. 
However, ad hoc modeling may incur additional planning time, as some algorithmic mechanism would have to decide when to apply estimation, effectively adding a computational overhead.
We denote
\begin{equation*}
\Delta_\textit{modeling}:=T_\textit{dynamic}^\textit{modeling}-T_\textit{offline}^\textit{modeling}, \Delta_\textit{planning}:=T_\textit{dynamic}^\textit{planning}-T_\textit{offline}^\textit{planning}
\end{equation*}
as the differences in running times of modeling and planning (not including run time of estimators) correspondingly due to using dynamic model construction.
Considering the combined run time of modeling and planning, we can determine that dynamic is computationally preferable to offline in case $|\Delta_\textit{modeling}| > |\Delta_\textit{planning}|$.

This criterion formally requires empirical testing per problem and planner, but may be practically useful when $T_\textit{dynamic}^\textit{modeling} \ll T_\textit{offline}^\textit{modeling}$ is projected, as it is reasonable to assume that $T_\textit{dynamic}^\textit{planning}$ and $T_\textit{offline}^\textit{planning}$ are on the same order of magnitude.
In conclusion, when $|\mathcal{A}_\textit{actual}| \ll n$ or $t_\textit{avg}^\textit{actual} \ll t_\textit{avg}^\textit{best}$ (or both) are projected, it is advantageous to turn to online modeling.

%% file: research_status.tex
\section{Dynamic Action Cost Estimation}

Our framework employs the basic assumption that every ground action can potentially have multiple cost estimators, with varying degrees of accuracy and different running times.
In particular we assume that once called, each estimator returns lower and upper bounds for the true action cost.
Note that this does not prevent knowledge of exact costs (where the bounds are simply equal), nor the usage of bound priors, that can be specified in the initial problem model (these can be thought of as estimators that have fast $O(1)$ run time).
It is worth mentioning that an anytime algorithm that serves as a cost estimator can in fact represent different estimators, where each of them is just an invocation of the same one but provided different running times. 

Relying on this assumption, we then define a deterministic planning problem where the goal of the planner is to find a plan that meets a target sub-optimality multiplier as fast as possible. I.e., it aims to efficiently find a plan $\pi^\epsilon$ that satisfies
\begin{equation*}
c(\pi^\epsilon) \le c^* \times \epsilon,
\end{equation*}
with $c^*$ being the optimal cost and~$\epsilon \ge 1$.
We proved that an algorithm which utilizes lower and upper bounds of costs, instead of exact values, can solve this problem by relying on the ratio of the accumulated bounds for the action costs composing the plan.

This lead to the development of \asec ($A^*$ with Synchronous Estimations of Costs) that implements this idea, by using the ratio of accumulated bounds to decide when to invoke cost estimation. \asec serves as our principal algorithm for solving such problem instances, and we have been able to prove that it is sound, and incomplete in general, but is complete under special circumstances.
We further developed a post-search procedure that builds on \asec to obtain an improved plan cost estimation when possible. 

We implemented \asec and its extensions by modifying and extending Fast Downward, and then empirically tested its performance on problems generated from planning competition benchmarks, which were added synthetic estimators.
Our findings provide strong empirical evidence that \asec outperforms alternatives w.r.t. run time, while typically meeting the target bound.
These results, along with detailed analysis, are summarized in another paper~\cite{weiss2022planning}.

%% file: challenges_future_work.tex
\section{Challenges and Future Work}
We suggest two possibilities for future research.
First, as mentioned earlier, the approach of online modeling includes two families: one that addresses the symbolic structure of an action model and another that is concerned with the numeric parameters of the model. In this paper we proposed an implementation that fixes the former and focuses on the latter.  Planning with online symbolic structure action models, represented in pure or mixed declarative form, that are updated during planning, is currently an entirely open challenge.

Second, considering problem representations other than classical planning, there are various numeric parameters characterizing action models, such as time duration and probability. Implementing our approach for these parameters constitutes another important direction for future work.


%% file: main.bbl
\begin{thebibliography}{18}
\providecommand{\natexlab}[1]{#1}

\bibitem[{Arora et~al.(2018)Arora, Fiorino, Pellier, M{\'e}tivier, and
  Pesty}]{arora2018review}
Arora, A.; Fiorino, H.; Pellier, D.; M{\'e}tivier, M.; and Pesty, S. 2018.
\newblock A review of learning planning action models.
\newblock \emph{The Knowledge Engineering Review}, 33.

\bibitem[{Baier et~al.(2008)Baier, Fritz, Bienvenu, and
  McIlraith}]{baier2008beyond}
Baier, J.~A.; Fritz, C.; Bienvenu, M.; and McIlraith, S.~A. 2008.
\newblock Beyond Classical Planning: Procedural Control Knowledge and
  Preferences in State-of-the-Art Planners.
\newblock In \emph{AAAI}, 1509--1512.

\bibitem[{Blythe(1999)}]{blythe1999decision}
Blythe, J. 1999.
\newblock Decision-theoretic planning.
\newblock \emph{AI magazine}, 20(2): 37--37.

\bibitem[{Boddy(2003)}]{boddy2003imperfect}
Boddy, M.~S. 2003.
\newblock Imperfect match: PDDL 2.1 and real applications.
\newblock \emph{Journal of Artificial Intelligence Research}, 20: 133--137.

\bibitem[{Dornhege et~al.(2009)Dornhege, Eyerich, Keller, Tr{\"u}g, Brenner,
  and Nebel}]{dornhege2009semantic}
Dornhege, C.; Eyerich, P.; Keller, T.; Tr{\"u}g, S.; Brenner, M.; and Nebel, B.
  2009.
\newblock Semantic attachments for domain-independent planning systems.
\newblock In \emph{Nineteenth International Conference on Automated Planning
  and Scheduling}.

\bibitem[{Eyerich(2013)}]{eyerich2013beyond}
Eyerich, P. 2013.
\newblock \emph{Beyond classical planning: temporal and probabilistic
  extensions}.
\newblock Ph.D. thesis, Dissertation, Universit{\"a}t Freiburg, 2013.

\bibitem[{Fox and Long(2003)}]{fox2003pddl2}
Fox, M.; and Long, D. 2003.
\newblock {PDDL} 2.1: An extension to {PDDL} for expressing temporal planning
  domains.
\newblock \emph{Journal of Artificial Intelligence Research}, 20: 61--124.

\bibitem[{Franc\`{e}s et~al.(2017)Franc\`{e}s, Rami\'{r}ez, Lipovetzky, and
  Geffner}]{frances17}
Franc\`{e}s, G.; Rami\'{r}ez, M.; Lipovetzky, N.; and Geffner, H. 2017.
\newblock Purely Declarative Action Descriptions are Overrated: Classical
  Planning with Simulators.
\newblock In \emph{Proceedings of the Twenty-Sixth International Joint
  Conference on Artificial Intelligence}.

\bibitem[{Garrett et~al.(2021)Garrett, Chitnis, Holladay, Kim, Silver,
  Kaelbling, and Lozano-P{\'e}rez}]{garrett2021integrated}
Garrett, C.~R.; Chitnis, R.; Holladay, R.; Kim, B.; Silver, T.; Kaelbling,
  L.~P.; and Lozano-P{\'e}rez, T. 2021.
\newblock Integrated task and motion planning.
\newblock \emph{Annual review of control, robotics, and autonomous systems}, 4:
  265--293.

\bibitem[{Ghallab, Nau, and Traverso(2016)}]{ghallab2016automated}
Ghallab, M.; Nau, D.; and Traverso, P. 2016.
\newblock \emph{Automated planning and acting}.
\newblock Cambridge University Press.

\bibitem[{Gnad et~al.(2019)Gnad, Torralba, Dom{\'\i}nguez, Areces, and
  Bustos}]{gnad2019learning}
Gnad, D.; Torralba, A.; Dom{\'\i}nguez, M.; Areces, C.; and Bustos, F. 2019.
\newblock Learning how to ground a plan--partial grounding in classical
  planning.
\newblock In \emph{Proceedings of the AAAI Conference on Artificial
  Intelligence}, 7602--7609.

\bibitem[{Helmert(2006)}]{helmert-jair2006}
Helmert, M. 2006.
\newblock The {Fast} {Downward} Planning System.
\newblock \emph{Journal of Artificial Intelligence Research}, 26: 191--246.

\bibitem[{Kambhampati(2007)}]{kambhampati2007model}
Kambhampati, S. 2007.
\newblock Model-lite planning for the web age masses: The challenges of
  planning with incomplete and evolving domain models.
\newblock In \emph{Proceedings of the National Conference on Artificial
  Intelligence}.

\bibitem[{Lipovetzky and Geffner(2017)}]{lipovetzky2017best}
Lipovetzky, N.; and Geffner, H. 2017.
\newblock Best-first width search: Exploration and exploitation in classical
  planning.
\newblock In \emph{Thirty-First AAAI Conference on Artificial Intelligence}.

\bibitem[{McCluskey(2003)}]{mccluskey2003pddl}
McCluskey, T.~L. 2003.
\newblock PDDL: A Language with a Purpose?
\newblock In \emph{ICAPS}.

\bibitem[{Rintanen(2015)}]{rintanen2015impact}
Rintanen, J. 2015.
\newblock Impact of modeling languages on the theory and practice in planning
  research.
\newblock In \emph{Twenty-Ninth AAAI Conference on Artificial Intelligence}.

\bibitem[{Weiss and Kaminka(2022)}]{weiss2022planning}
Weiss, E.; and Kaminka, G.~A. 2022.
\newblock Planning with Dynamically Estimated Action Costs.
\newblock In \emph{Proceedings of the 1st ICAPS Workshop on Reliable
  Data-Driven Planning and Scheduling}.

\bibitem[{Weyhrauch(1980)}]{weyhrauch1980prolegomena}
Weyhrauch, R.~W. 1980.
\newblock Prolegomena to a theory of mechanized formal reasoning.
\newblock \emph{Artificial intelligence}, 13(1-2): 133--170.

\end{thebibliography}
